\documentclass{article}

 \usepackage[preprint]{neurips_2026}

\usepackage[utf8]{inputenc} 
\usepackage[T1]{fontenc}    
\usepackage{hyperref}       
\usepackage{url}            
\usepackage{booktabs}       
\usepackage{amsfonts}       
\usepackage{nicefrac}       
\usepackage{microtype}      
\usepackage[table]{xcolor}         
\usepackage{amsmath}
\usepackage{graphicx}
\usepackage{subcaption}
\usepackage{makecell}
\usepackage{tabularx}
\usepackage{multirow}
\usepackage{pifont}
\usepackage{xcolor}
\usepackage{colortbl}
\usepackage{wrapfig}

\definecolor{mygray}{gray}{0.85}
\title{HiFloat4 Format for Language Model Pre-training on Ascend NPUs}

\author{
Mehran Taghian \quad Yunke Peng \quad Xing Huang \quad Yao Wang \quad Yaoyuan Wang \quad Wei Guo \\
Yuanyong Luo \quad Tianchi Hu \quad Junsong Wang \quad Xin Wang \quad Hu Liu \quad Yu Cheng \quad Ziwei Yu \\
Hongliang Li \quad Mehdi Rahimifar \quad Lei Yan \quad Xuefei Wang \quad Zhuang Ma \quad Lei Liu \quad Hui Yu \\
Anandharaju Durai Raju \quad Hoang Le \quad Hei Yi Mak \quad Tanzila Rahman \quad Shadan Golestan \\
\vspace{0.2cm} \\
\textbf{Huawei} \\
\vspace{0.2cm} \\
\texttt{\{mehran.taghian1, pengyunke\}@huawei.com}
}

\begin{document}

\maketitle

\begin{abstract}
Training large foundation models at low numerical precision is one of the most
promising directions for reducing the compute and memory cost of modern AI.
Recent 4-bit floating-point formats such as MXFP4 and NVFP4 can be applied to
linear GEMM operations in LLMs, but their limited dynamic range introduces
numerical instability that prior work addresses by stacking stabilization
mechanisms, typically executed at higher precision and partially eroding the
efficiency gains that motivate FP4. In this work, we argue that numerical format
design is itself a first-class lever for stable FP4 training, and present the
first systematic study of FP4 LLM pretraining on energy-efficient Huawei Ascend
NPUs. We compare the recently proposed HiFloat4 (HiF4) format against
both MXFP4 and NVFP4, holding one recipe fixed across all three
formats across dense (OpenPangu-1B, Llama3-8B) and Mixture-of-Experts
(Qwen3-MoE-30B) architectures and executing all linear and expert GEMMs in FP4.
At matched storage --- NVFP4 and HiF4 both spend 4.5 bits per value --- the
three formats differ far more in what they require before they will train at all
than in final accuracy. HiF4 reaches a relative loss of 1.55\% with no
stabilization, below fully stabilized MXFP4 (1.79\%) and below NVFP4 carrying the per-tensor scaling it cannot train without (2.00\%); NVFP4 diverges under every combination of stochastic rounding and Hadamard transform we tried. Our results suggest that stable, accurate FP4 training does not require an ever-growing stack of stabilization techniques; it requires the right numerical format.
\end{abstract}

\section{Introduction}
The computational cost of training large language models (LLMs) has become a critical bottleneck. Empirical scaling laws show that performance improves as a power-law function of model size, data, and compute \citep{kaplan2020scaling}, making it essential to reduce per-step cost so capacity and data can scale without prohibitive hardware and energy demands. Low-precision training is among the most promising directions: representing parameters, activations, and gradients at reduced precision directly cuts memory bandwidth and compute. Prior work has shown that 8-bit floating-point formats can train LLMs effectively when paired with fine-grained block scaling \citep{micikevicius2022fp8, liu2024deepseek}, and recent efforts have pushed this frontier to 4-bit formats such as MXFP4 \citep{rouhani2023mxfp4} and NVFP4 \citep{abecassis2025pretraining}. Yet FP4's limited representational capacity introduces numerical instability that can derail optimization entirely.

A central source of instability is systematic bias in gradient quantization. When gradients contain large outliers, block-level scaling factors are dominated by peak magnitudes, leaving most values confined to a narrow portion of the representable range — and many to underflow to zero — disrupting backpropagation. To compensate, recent work has stacked stabilization techniques: Stochastic Rounding (SR)~\citep{croci2022stochastic}, Random Hadamard Transforms (RHT) for gradient decorrelation \citep{tseng2025training, abecassis2025pretraining}, engineered block-scaling schemes, and other format-specific mechanisms \citep{cao2025metis, castro2025quartet, panferov2026quartet, cook2025four}. These auxiliary operations typically run at higher precision (e.g., FP16 or FP32) before FP4 quantization, partially offsetting the efficiency gains that motivated low-precision training. This raises a natural question: rather than patching a format's limitations with increasingly complex stabilization pipelines, can a better-designed numerical representation reduce this burden from the outset?

We argue that numerical format design is itself a first-class lever for stable and efficient FP4 training, and investigate LLM pretraining using the HiFloat4 (HiF4) format \citep{luo2026hifloat4}, a 4-bit representation targeting next-generation AI accelerators. HiF4 employs a three-level hierarchical scaling scheme over blocks of 64 elements: an E6M2 top-level scaler, E1 intermediate scalers, and an S1P2 (equivalently E1M2) representation for the FP4 values. By distributing mantissa bits across both global scalers and local elements, this design expands dynamic range while preserving precision — substantially reducing the need for auxiliary stabilization. We evaluate HiF4 against the two FP4 formats with announced hardware support, MXFP4
\citep{rouhani2023mxfp4} and NVFP4 \citep{abecassis2025pretraining}. NVFP4 is the necessary storage-matched control: its 8-bit E4M3 block scale is amortized over 16 elements, so it costs 4.5 bits per value exactly as HiF4 does, where MXFP4 costs 4.25. At matched storage the accuracy ordering tightens considerably --- NVFP4 closes most of the distance between MXFP4 and HiF4, and the margin that remains is 0.08 to 0.14 points, which we do not lean on. What separates the formats is what each needs before it trains at all. HiF4 with no stabilization reaches a lower relative loss than MXFP4 carrying all three of its stabilizers and than NVFP4 carrying the per-tensor scaling (PTS) without which it diverges, under every combination of SR and RHT we tried.

Beyond format comparison, our broader goal is efficient FP4 pretraining on energy-constrained accelerators. We target Huawei Ascend NPUs — domain-specific accelerators with specialized matrix, vector, and scalar units that offer improved power efficiency over general-purpose GPUs — and develop an FP4 training pipeline that executes the majority of linear operations in FP4 while maintaining stable optimization. All experiments run on Ascend hardware, but not on FP4 kernels: no Ascend NPU supports HiF4 or MXFP4 natively yet, so every GEMM executes in BF16 with quantization simulated around it, bit-exactly. We therefore measure the numerical behaviour native kernels would produce, but no throughput, latency or energy. On that basis we evaluate both dense Transformers (OpenPangu-1B, Llama3-8B) and a large Mixture-of-Experts (MoE) architecture (Qwen3-MoE-30B), with $\geq75\%$ of training computation in simulated FP4 and a loss gap within $\approx$1\% of a full-precision baseline.

Our main contributions are:
\begin{itemize}
    \item We argue that numerical format design is itself a first-class mechanism for stable FP4 training, and that a well-designed representation can reduce the reliance on auxiliary high-precision stabilization pipelines that erode the efficiency gains of low-precision training.

    \item We investigate HiF4 for pretraining, which under matched conditions, attains a relative loss gap within 1\% using only a single stabilization technique—roughly half the gap of MXFP4, which requires multiple stabilization mechanisms to reach comparable stability—while sustaining less than 1\% average degradation across downstream tasks.

    \item We hold one training recipe fixed across three FP4 formats and find that they differ primarily in stabilization dependence rather than in final accuracy: HiF4 is competitive with no stabilizer at all, MXFP4 needs three, and NVFP4 does not converge without PTS. We do not claim the format itself, which was introduced for inference; what is new is that its hierarchical scaling suffices for stable pretraining, where gradient quantization is the hard part.

    \item We develop an end-to-end FP4 pretraining pipeline targeting Huawei Ascend NPUs that executes the majority of linear operations in FP4 while maintaining stable optimization on domain-specific, energy-efficient accelerators.

    \item We validate the pipeline on Dense and MoE transformer models, running $\geq 75\%$ of training computation in FP4 with a loss gap within ${\approx}1\%$ of full-precision baselines.
\end{itemize}

To the best of our knowledge, this is the first systematic study of FP4 LLM pretraining on Ascend NPUs. More broadly, our results suggest that stable, accurate FP4 training does not require an ever-growing stack of stabilization techniques — it requires the right numerical format. HiF4's hierarchical scaling provides enough representational headroom to achieve low relative error on its own; a single addition — RHT to reduce noise in $dW$ — closes the remaining gap, yielding a pipeline that is both simpler and more accurate than prior FP4 approaches. A corollary is that stabilization recipes belong to a format and do not port across formats: SR is necessary for MXFP4, actively harmful for HiF4, and insufficient for NVFP4 at any setting.

\section{Related Works}

\textbf{General FP4 Training} 
Several works study 4-bit training without targeting a specific hardware format.
LUQ \citep{chmiel2021accurate} pairs INT4 weights/activations with FP4 gradients
and introduces stochastic underflow to debias subnormal gradients.
\cite{zhou2025towards} applies per-block FP4 (block size 128) to FFN forward
GEMMs only, keeping attention and backward GEMMs in FP8---a mixed-precision
rather than end-to-end FP4 scheme.
\cite{hu2025elucidating} identifies principles for stable FP4 training: RHT for
outlier redistribution, tensor scaling, SR, and a UE5M3 update-scale format
(block size 16).
\cite{chen2025int} develops a QSNR theory for block-wise quantization, showing
INT can outperform FP depending on crest factor and bit width.

\textbf{Handling Outliers}
A complementary line of work attributes low-bit degradation primarily to
outliers. \cite{zhang2025accurate} detects outlier-heavy blocks and falls back
to higher precision past a threshold, while \cite{wang2025optimizing} clamps
activation outliers and processes the sparse residual in higher precision.
Such fallback mechanisms add compute and memory overhead, cause load imbalance,
and limit effective FP4 hardware utilization.

\textbf{Training with MXFP4}
Naively applying MXFP4 to all GEMMs is unstable, motivating several stabilization
recipes. Tetrajet \citep{chen2025oscillation} attributes degradation to biased
gradients and forward weight oscillations near quantization thresholds, and
addresses them with truncation-free scaling, unbiased double quantization, SR,
and the Q-EMA/Q-Ramping stabilizers---at the cost of extra compute and EMA state.
Quartet \citep{castro2025quartet} decouples forward and backward objectives:
Hadamard rotations with RMSE-based clipping (QuEST) preserve effective model
capacity in the forward, and SR with RHT preserves effective data efficiency in
the backward. \cite{tseng2025training} closes the BF16--MXFP4 gap using SR,
RHT, and truncation-free scaling, but quantizes only backward GEMMs; in contrast,
we quantize both forward and backward GEMMs and apply RHT selectively to the
weight-gradient computation. M2XFP \citep{hu2026m2xfp} augments MXFP with
element-level (activations) and subgroup-level (weights) metadata to preserve
extreme values, at the cost of additional storage and metadata management.

\textbf{Training with NVFP4}
NVFP4 is NVIDIA's Blackwell-supported FP4 format and requires algorithmic support for stable training. \cite{cook2025four} proposes Four-over-Six, mapping block maxima to 4 or 6 to reduce rounding error near the FP4 maximum, at the cost of dual quantization passes. Quartet II \citep{panferov2026quartet} introduces MS-EDEN, which combines RHT with SR applied to micro-scale factors rather than individual FP4 values, and uses Nearest Rounding (NR) in the forward and MS-EDEN in the backward to obtain unbiased gradients. NVIDIA's recipe \citep{abecassis2025pretraining} shows that stable FP4 pretraining at scale requires a combination of selective high-precision layers, SR, RHT, 2D weight quantization for forward--backward consistency, and optional late-stage precision switching. \cite{chen2025power} improves NVFP4 efficiency by repurposing the redundant negative zero and sign bit of the FP8 scaling factor to encode additional quantization information.

\textbf{MXFP4 vs.\ NVFP4 Comparisons}
\cite{chmiel2025fp4} shows that fully quantized FP4 LLM training is feasible
at scale by combining NVFP4 microscaling (block size 16, E4M3 scale) with
selective SR on gradients and backward activations while keeping NR in the
forward for stability. Metis \citep{cao2025metis} argues that spectral
anisotropy is the dominant obstacle to low-bit training and uses spectral
decomposition, adaptive spectral learning rates, and dual-range regularization
to match full-precision performance under both MXFP4 and NVFP4.

\textbf{MoE Models}
Low-precision MoE training is complicated by routing dynamics and sparse expert
activation. \cite{zhang2026practical} stores and communicates activations in
FP4 but computes in FP8 to maintain stability.
\cite{yan2026scalabletrainingmixtureofexpertsmodels} integrates the techniques
of \cite{abecassis2025pretraining} into a scalable MoE training framework, but
only focuses on system throughput.

\textbf{Our Work}
Our work differs from prior FP4 pretraining efforts in three ways. First, where existing studies target general-purpose GPUs, we present the first systematic study of FP4 LLM pretraining on energy-efficient NPU hardware (Huawei Ascend). Second, rather than stacking stabilization mechanisms — SR, RHT, and format-specific scaling schemes, typically executed at higher precision and eroding the very efficiency gains that motivate FP4 — we treat numerical format design as a first-class lever, building on HiF4 \citep{luo2026hifloat4}, whose hierarchical scaling provides enough representational headroom that a single stabilization step (RHT on $dW$) suffices to close the gap to full precision. Third, we validate our approach on both Dense and MoE architectures, executing $\geq75\%$ of training in FP4 within $\approx$1\% of a full-precision baseline. Finally, where the works above evaluate NVFP4 in isolation or against MXFP4, we train it end-to-end under the same pipeline, models, token budget and stabilizer sweep as HiF4 and MXFP4, making it a storage-matched control rather than a cited number.
\section{Low-Precision FP4 Training}
\label{sec:method}
\begin{figure}[ht]
    \centering
    \begin{subfigure}[b]{0.228\textwidth}
        \centering
        \includegraphics[width=\textwidth]{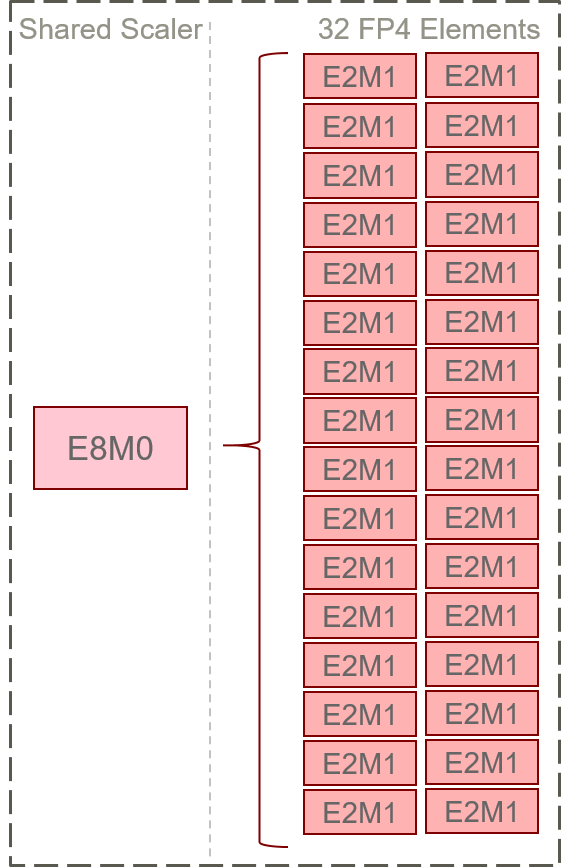}
        \caption{MXFP4}
        \label{fig:mxfp4}
    \end{subfigure}
    \hfill
    \begin{subfigure}[b]{0.195\textwidth}
        \centering
        \includegraphics[width=\textwidth]{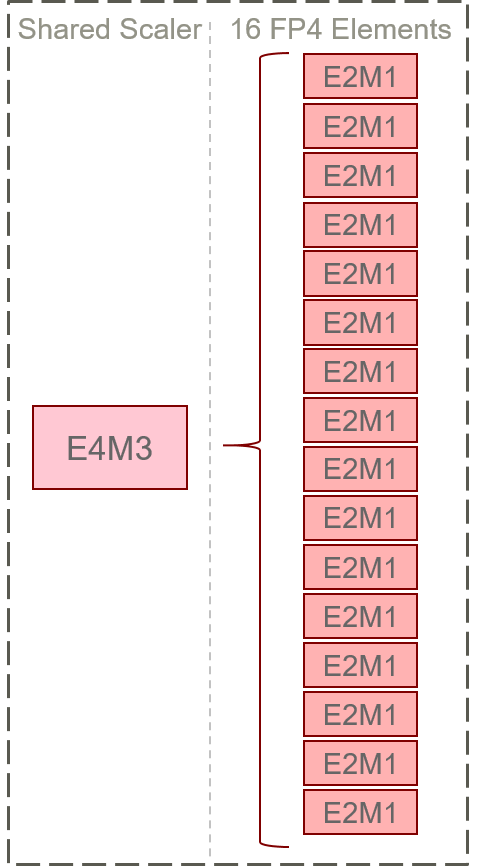}
        \caption{NVFP4}
        \label{fig:nvfp4}
    \end{subfigure}
    \hfill
    \begin{subfigure}[b]{0.56\textwidth}
        \centering
        \includegraphics[width=\textwidth]{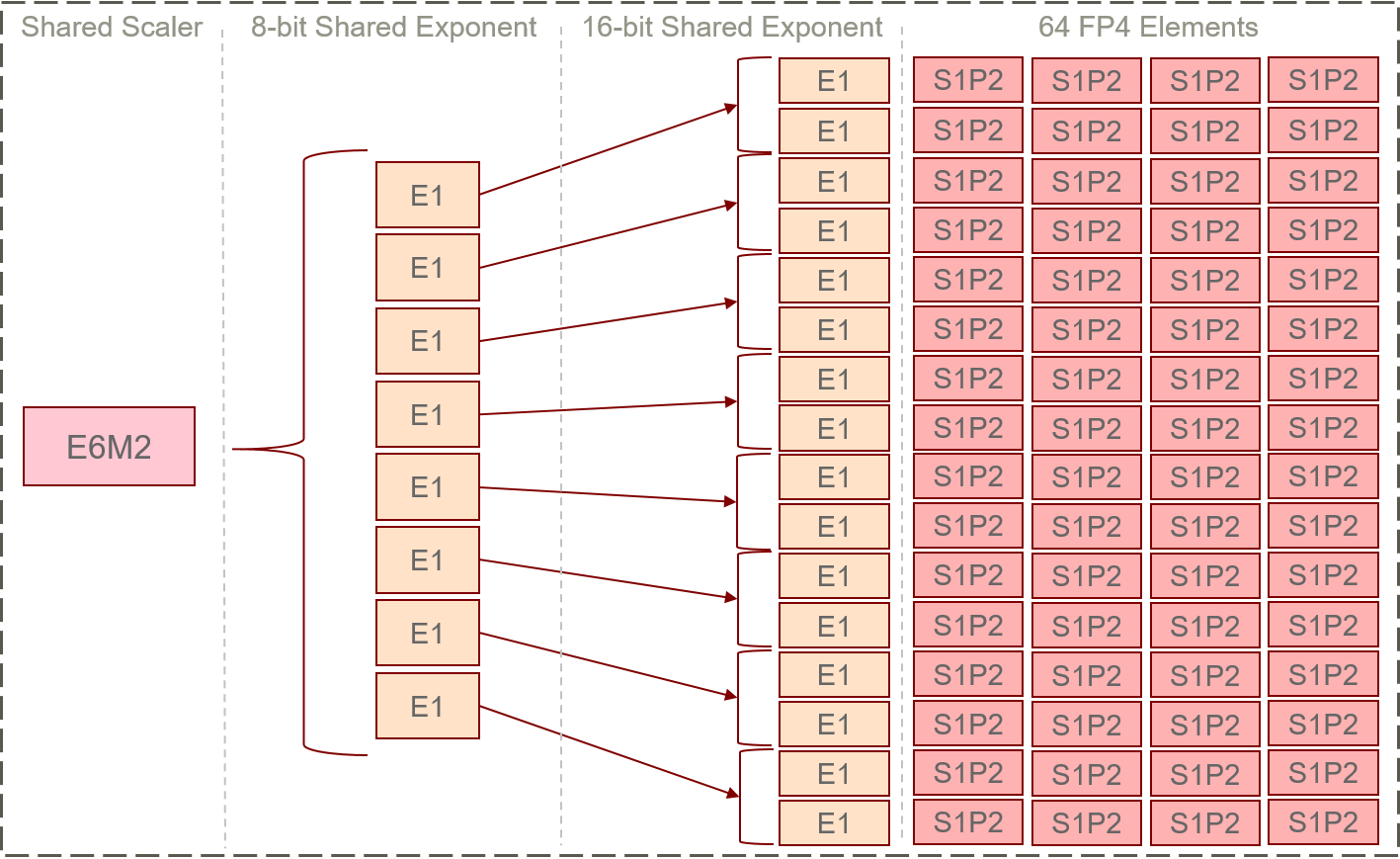}
        \caption{HiFloat4}
        \label{fig:hif4}
    \end{subfigure}
    \caption{FP4 format architectures. (a) MXFP4: 8 bits of shared metadata (E8M0) over 32 4-bit elements, 4.25 bits per value. (b) NVFP4: 8 bits (E4M3) over 16 elements, 4.5 bits per value, plus an FP32 per-tensor scale. (c) HiF4: 32 bits of three-level metadata over 64 elements, also 4.5 bits per value. NVFP4 and HiF4 are therefore storage-matched; MXFP4 is cheaper but coarser.}
    \label{fig:fp4-arch}
\end{figure}
We employ FP4 computation across all linear layers, including both Dense and MoE components. Specifically, we adopt the recently proposed MXFP4, NVFP4 and HiF4 formats. MXFP4 and HiF4 are candidate formats for next-generation Ascend NPUs and NVFP4 is the storage-matched control (Section~\ref{sec:npu}); no FP4 arithmetic in this paper runs natively on any of them, and all three are emulated (Section~\ref{sec:experiments}). All three are used to enable end-to-end FP4 training of large language models (LLMs). MXFP4 \citep{rouhani2023microscaling, rouhani2023mxfp4} is a 4-bit format using block-wise microscaling: each block of 32 elements is encoded in E2M1 with a shared E8M0 scale. Figure~\ref{fig:mxfp4} illustrates the architecture. NVFP4 \citep{abecassis2025pretraining} uses the same E2M1 elements as MXFP4 but a smaller block of 16 with a shared E4M3 scale, layered under a single FP32 per-tensor scale (Figure~\ref{fig:nvfp4}). Amortizing 8 bits of block scale over 16 rather than 32 elements costs 4.5 bits per value --- identical to HiF4, above MXFP4's 4.25 --- which makes NVFP4 the storage-matched comparison for HiF4 and MXFP4 the cheaper but coarser one. HiF4 is a hardware-oriented FP4 format for Huawei Ascend NPUs that employs three-level hierarchical scaling. A shared unsigned E6M2 factor provides a coarse block-level scale, while two finer levels of 1-bit micro-exponents refine 8- and 16-way partitions, with each level-3 exponent governing four contiguous FP4 elements. Data elements use S1P2 encoding (equivalent to E1M2). Figure~\ref{fig:hif4} illustrates the architecture. See Appendix~\ref{sec:low-precision-formats} for full details. 

\begin{table}[t]
\centering

\caption{Extra compute beyond FP4 GeMM and quantization for methods that apply FP4 in both forward/backward during pretraining.
$X\in\mathbb{R}^{M\times K}$, $W\in\mathbb{R}^{N\times K}$, and $D\in\mathbb{R}^{M\times N}$. 
$r$ is the RHT block size; $k, j$ are Metis's forward/backward randomized SVD ranks. Quartet II's 4/6 forward scale selection adds a $\Theta(MK+NK)$ constant-factor overhead to forward quantization, excluded here. ``--'' indicates no additional compute beyond baseline GeMM and quantization. The table isolates stabilization cost and excludes the baseline FP4 GEMM and quantization, which
every method pays; Quartet II's 4/6 selection is a constant-factor change to double quantization, which belongs to that baseline path, so it falls inside the exclusion as a scoping decision. See Appendix~\ref{sec:compute-analysis}.}
{\everymath{\scriptstyle}
\begin{tabularx}{\textwidth}{lccc}
\toprule
 & Forward & \multicolumn{2}{c}{Backward} \\
\cmidrule(lr){2-2} \cmidrule(lr){3-4}
Method & $Y = XW^\top$ & $dX = DW$ & $dW = D^\top X$ \\
\midrule
NVIDIA~\citep{abecassis2025pretraining}
  & --
  & --
  & $(N+K)M \log r$ \\
Quartet II~\citep{panferov2026quartet}
  & --
  & $(M+K)N \log r$
  & $(N+K)M \log r$ \\
Metis~\citep{cao2025metis}
  & $Mk(K+N)$
  & $MNj + jk(M+N+K)$
  & $Kj(M+N) + jk(M+N+K)$ \\
\bottomrule
\end{tabularx}}
\label{tab:compute}
\end{table}

Stabilizing training under such extreme quantization demands additional algorithmic techniques beyond standard low-precision arithmetic. To this end, we analyze three state-of-the-art FP4 training frameworks in Table~\ref{tab:compute}, with a focus on their auxiliary high-precision computational overhead beyond FP4 quantization and GEMM operations. For details of how we analyze the computation complexities, see Appendix~\ref{sec:compute-analysis}. Such auxiliary computations can become a significant bottleneck at scale, diminishing the effective speedup otherwise afforded by low-precision training. Among the three approaches considered—Quartet II \citep{panferov2026quartet}, Metis \citep{cao2025metis}, and the method proposed by NVIDIA \citep{abecassis2025pretraining}—the last incurs the lowest computational overhead. We adopt it as our baseline for FP4 training, incorporating the following modifications:
\begin{enumerate}
    \item 2D weight quantization is omitted. We make no claim about what it would recover and do not estimate its hardware cost; the omission applies identically to all three formats, and we record it in the limitations as an untested deviation from the NVIDIA recipe.
    \item Randomized Hadamard Transform (RHT) is applied to the $dW$ computation for both MXFP4 and HiF4. Additionally, for MXFP4, we apply SR to the gradient tensor, together with the truncation-free scaling strategy of TetraJet, which adjusts the block scale to keep all values in range rather than clipping out-of-range magnitudes. For NVFP4 we apply SR to the gradient tensor on top of PTS.
    \item FP4 training is applied uniformly across all linear layers (except the output and embedding layers), without retaining high-precision computation for selectively sensitive components.
\end{enumerate}

{\textbf{Per-tensor scaling (PTS).} NVFP4's E4M3 block scale spans a far narrower exponent range than MXFP4's E8M0 or HiF4's E6M2 plus micro-exponents --- 22 binades against HiF4's 69 --- so blocks near either edge of it saturate or underflow. PTS corrects this with a single FP32 scale per tensor, derived from the tensor's global maximum and applied before micro-scaling, so that the per-block E4M3 factors encode only variation \emph{within} the tensor. PTS is part of NVFP4's training formulation rather than an optional addition. We nonetheless count it as a stabilization technique throughout this paper, because what we are measuring is the auxiliary high-precision work each format requires outside the FP4 kernel. PTS is a full-tensor reduction and rescale performed in BF16 before quantization, which is the same cost SR, RHT and TF scaling impose. Classifying it by where it runs, rather than by whether the format's authors regard it as optional, is what keeps Table~\ref{tab:compute} and Section~\ref{sec:ablation} consistent with each other and with how we account for the other two
formats.


A primary difficulty in extremely low-bit training is how scaled elements are mapped to representable values. NR maps each element to its closest representable value; while hardware-friendly, it introduces a systematic bias that is particularly harmful for gradients, accumulating over iterations. SR introduces controlled randomness prior to rounding and is unbiased in expectation. SR is not uniformly beneficial across formats, however. HiF4's higher per-element mantissa bits and global scaler substantially reduce quantization bias even without SR, so the variance SR introduces outweighs its bias-reduction benefit and degrades performance relative to plain NR. MXFP4 operates at coarser precision, where the bias is more pronounced and SR provides a meaningful correction. As we demonstrate empirically in Section~\ref{sec:ablation}, SR improves training performance under MXFP4 but is detrimental under HiF4. NVFP4 falls on the MXFP4 side of this divide: it shares the E2M1 element encoding and so the same coarse per-element precision, and SR yields a net benefit once PTS is in place --- though SR does not by itself make the format trainable (Section~\ref{sec:ablation}).

We also apply an RHT to inputs prior to quantization. RHT redistributes energy more uniformly across dimensions, suppressing outliers and reducing the per-coordinate dynamic range; this mitigates underflow of small values in the presence of large outliers and, with SR, reduces the variance of the injected noise. See Appendix~\ref{sec:stabilization}.

\begin{figure}
    \centering
    \includegraphics[width=1\linewidth]{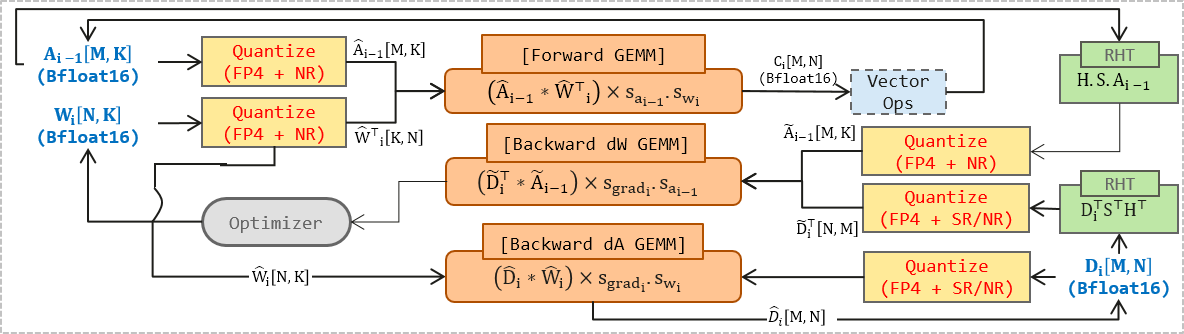}
    \caption{Overview of the GEMM operation within a linear or expert layer. Activations and weights are quantized to HiF4, MXFP4 or NVFP4 using NR; gradients use SR for MXFP4 and NVFP4 and NR for HiF4. An RHT is applied to the weight gradients ($\mathrm{d}W$) prior to quantization. The flow is identical for all three formats; only the quantizer and its stabilizers change.}
    \label{fig:gemm}
\end{figure}
As illustrated in Figure~\ref{fig:gemm}, a linear layer involves three GEMM operations: 1) Forward pass $A_{i} = \phi(A_{i-1} W_i^\top)$, 2) Backward pass (gradient w.r.t. weight) $dW_i = grad_i^\top A_{i-1}$, 3) Backward pass (gradient w.r.t. input) $dA_{i-1} = grad_i W_{i}$,
where $W_i$ denotes the weight matrix of layer $i$, $A_{i-1}$ is the input activation from layer $i-1$, $\mathrm{grad}_i$ is the upstream gradient propagated from layer $i+1$, and $\phi(\cdot)$ represents the activation function.
In the forward GEMM, both activations and weights are quantized to FP4 using NR. During the backward pass, gradient tensors are quantized to FP4 using either NR or SR to mitigate bias accumulation. Empirically, we observe that SR improves training stability and performance for MXFP4, whereas HiF4 achieves better results when gradients are quantized using NR.
Furthermore, for the computation of $\mathrm{d}W_i$, RHT with a block size of $k=64$ is applied along the inner dimension prior to quantization. The transformed tensor is then quantized to MXFP4/HiF4, improving numerical stability under low-precision accumulation.
The overall workflow of the quantized linear layer is summarized in Figure~\ref{fig:gemm}.
\section{4-bit Floating Point on Ascend NPUs}
\label{sec:npu}

Huawei Ascend NPUs \citep{huawei2023huawei} are widely used for large-scale LLM training and inference. Their compute is dominated by the Cube Unit, an array of processing elements (PEs) that perform parallel dot-product and accumulation operations. With a 256-bit input bandwidth, each PE can process a full HiF4 block of 64 FP4 elements in a single dot-product, aligning the HiF4 block size with the PE datapath and minimizing scaling and intermediate reductions. Two properties separate HiF4 from NVFP4 on this datapath. First, one HiF4 block matches the 64-length dot product of a PE, so a single unit pair suffices where NVFP4's 16-element group needs four. Second, applying a scale costs more for NVFP4: HiF4's E6M2 top-level factor needs a floating-point
multiply once per 64 elements, its level-2 and level-3 factors reducing to shifts, whereas NVFP4's E4M3 block scale needs one per 16 elements, plus the per-tensor factor PTS introduces. \citet{luo2026hifloat4} quantify the consequence at six fewer multipliers, roughly a third of the incremental area, and about 10\% less power. We need to emphasize that we measure no area, power or throughput anywhere in this paper. We make no equivalent claim against MXFP4, whose power-of-two E8M0 scale requires no floating-point multiplier at all and which is likewise well matched to the Ascend datapath.
\section{Experimental Analysis}
\label{sec:experiments}
Our experiments aim to quantify the FP4 and BF16 pretraining gap. Concretely, we investigate:
\begin{enumerate}
    \item \textit{Pretraining dynamics.} We compare loss curves under BF16, HiF4, and MXFP4, and measure their relative deviation throughout training.
    \item \textit{Downstream quality.} We evaluate pretrained models on downstream tasks to assess how low-precision training affects downstream performance.
    \item \textit{Sensitivity analysis.} We ablate the design choices that maintain stability and accuracy in the low-precision regime.
\end{enumerate}
\textbf{Emulation.} No FP4 GEMM in this paper runs natively. All GEMMs execute in BF16 on Ascend
910B3 NPUs with quantization and dequantization emulated around them at bit-exact precision, so the loss
curves are what native hardware would produce for the same recipe; what is unmeasurable here is speedup,
not accuracy, and we report no throughput, latency or energy figures. All runs share hardware, software,
seeds and data order and differ only in the quantization path (Appendix~\ref{sec:emulation}). Training
is performed on a cluster of 64 nodes, each equipped with 8 NPUs (65 GB of memory per NPU), 192 CPU
cores, and 1536 GB of host memory.

\textbf{Models.} We study three architectures: two dense Transformers, OpenPangu-1B and Llama3-8B, and one mixture-of-experts (MoE) model, Qwen3-MoE-30B. In the dense models, linear layers reside in the attention block (QKV and output projections) and the FFN (FC1 and FC2). These linear components account for $\approx 80\%$ of the parameters in OpenPangu-1B and $\approx 77\%$ in Llama3-8B. For Qwen3-MoE-30B, we apply FP4 to the QKV and output projections together with the up- and down-projections inside each expert. The attention projections contribute only $\approx 5\%$ of total compute, whereas the experts hold $\approx 90\%$ of model parameters.

\textbf{FP4 coverage.} Table~\ref{tab:layers} reports the fraction of compute executed in FP4 for each model component. FP4 is applied uniformly to every linear layer except output and embedding, for all three formats. This is stricter than NVIDIA's recipe, which keeps roughly 15\% of linear layers in higher precision and reports divergence when every one is quantized \citep{abecassis2025pretraining}, so our NVFP4 results are measured under a harsher regime than that work uses. 
Low-precision training is particularly advantageous for MoE: although the experts hold $\approx 90\%$ of parameters, top-8 routing over 128 experts activates only $6.25\%$ of experts per forward/backward pass. As a result, only $\approx 16\%$ of the model's parameters participate in computation at any given step, of which $\approx 75\%$ reside in linear layers and are executed in FP4. The remaining $\approx 84\%$ of parameters, corresponding to inactive experts, can potentially be \emph{stored} in FP4 without incurring any additional compute. FP4 thus delivers a double benefit on MoE models: memory savings on the dormant majority of weights, and throughput gains on the active subset.

\textbf{Training setup.} We use a unified optimizer configuration across all models: 50B tokens with Adam under a cosine learning-rate schedule, global batch size 2048 for OpenPangu and 512 for Llama and Qwen, with 480 and 2{,}000 warmup steps respectively. Full hyperparameters are in Appendix~\ref{sec:config}.

\begin{table}[t]
    \centering
    \caption{Percentage of compute executed in FP4, broken down by component}
    \begin{tabular}{lcccccc|c}
        \toprule
         & \multicolumn{2}{c}{Attention} & \multicolumn{2}{c}{FFN} & \multicolumn{2}{c}{Experts} &  \\
        \cmidrule(lr){2-3} \cmidrule(lr){4-5} \cmidrule(lr){6-7}
        Model & QKV & Output & FC1 & FC2 & Up & Down & Total \\
        \midrule
        OpenPangu-1B & 10.61\% & 5.30\% & 42.45\% & 21.22\% & -- & -- & 79.58\% \\
        Llama3-8B & 8.86\% & 5.91\% & 41.37\% & 20.68\% & -- & -- & 76.82 \% \\
        Qwen3-MoE-30B & 3.13\% & 2.50\% & -- & -- & 60.18\% & 30.09\% & 95.90\% \\
        \bottomrule
    \end{tabular}
    \label{tab:layers}
\end{table}
\subsection{Pretraining Dynamics}
Figure~\ref{fig:fp4-curves} compares the three FP4 schemes against the BF16 baseline, each in its own best configuration. HiF4 tracks BF16 at least as closely as the best-tuned MXFP4 and NVFP4 recipes while requiring the fewest stabilization techniques: HiF4 uses only RHT, MXFP4 additionally requires TF-scaling and SR, and NVFP4 additionally requires SR on top of the PTS. We attribute this to the higher per-element mantissa precision of HiF4 and its
hierarchical scaling, which reduces the need for auxiliary stabilization the other two formats require to recover representational capacity.

To quantify the gap, we compute the relative loss error between each low-precision run and the BF16 baseline:
$\text{E}_\text{Relative} = (\text{loss}_\text{low-precision} -\text{loss}_\text{baseline})/{\text{loss}_\text{baseline}}$.
Table~\ref{tab:relative-error} reports this quantity for the curves in Figure~\ref{fig:fp4-curves}.
HiF4 achieves lower relative error than both NVFP4 and MXFP4, and on Llama3-8B and Qwen3-MoE-30B its gap to BF16 falls below 1\%. The comparison against NVFP4 is much tighter than against MXFP4. Ordered by active parameter count rather than total parameters — OpenPangu-1B (1B), Qwen3-MoE-30B (4.6B), Llama3-8B (8B) — the HiF4 relative error decreases monotonically from 1.19\% to 0.88\% to 0.85\%, while its advantage over NVFP4 grows from 0.01 to 0.11 to 0.14 points. The gap between the two formats therefore widens as model scale increases.
We note that this comparison treats dense and MoE models on the same axis; total MoE parameters may still contribute to representational capacity in ways that active count alone does not capture.

The Llama3-8B HiF4 run (Figure~\ref{fig:fp4-curves}, top middle) shows two visible loss spikes, and the 0.85\% we report is from that run rather than a clean rerun. Both occur before roughly 10B tokens; thereafter the loss rejoins the BF16 trajectory for the remaining 40B. Spikes of this kind appear in all our runs, the BF16 reference included; HiF4's are the most visible but recover rather than diverge.
The MXFP4 configuration on the same model shows fewer spikes, indicating that the additional TF-scaling and SR stabilizers dampen transient instabilities, at the cost of higher steady-state loss and per-step overhead. Section~\ref{sec:ablation} presents an ablation study on FP4 stabilization methods, showing that the configuration used here is the most effective.

\begin{figure}[!htbp]
    \centering
    \includegraphics[width=\textwidth]{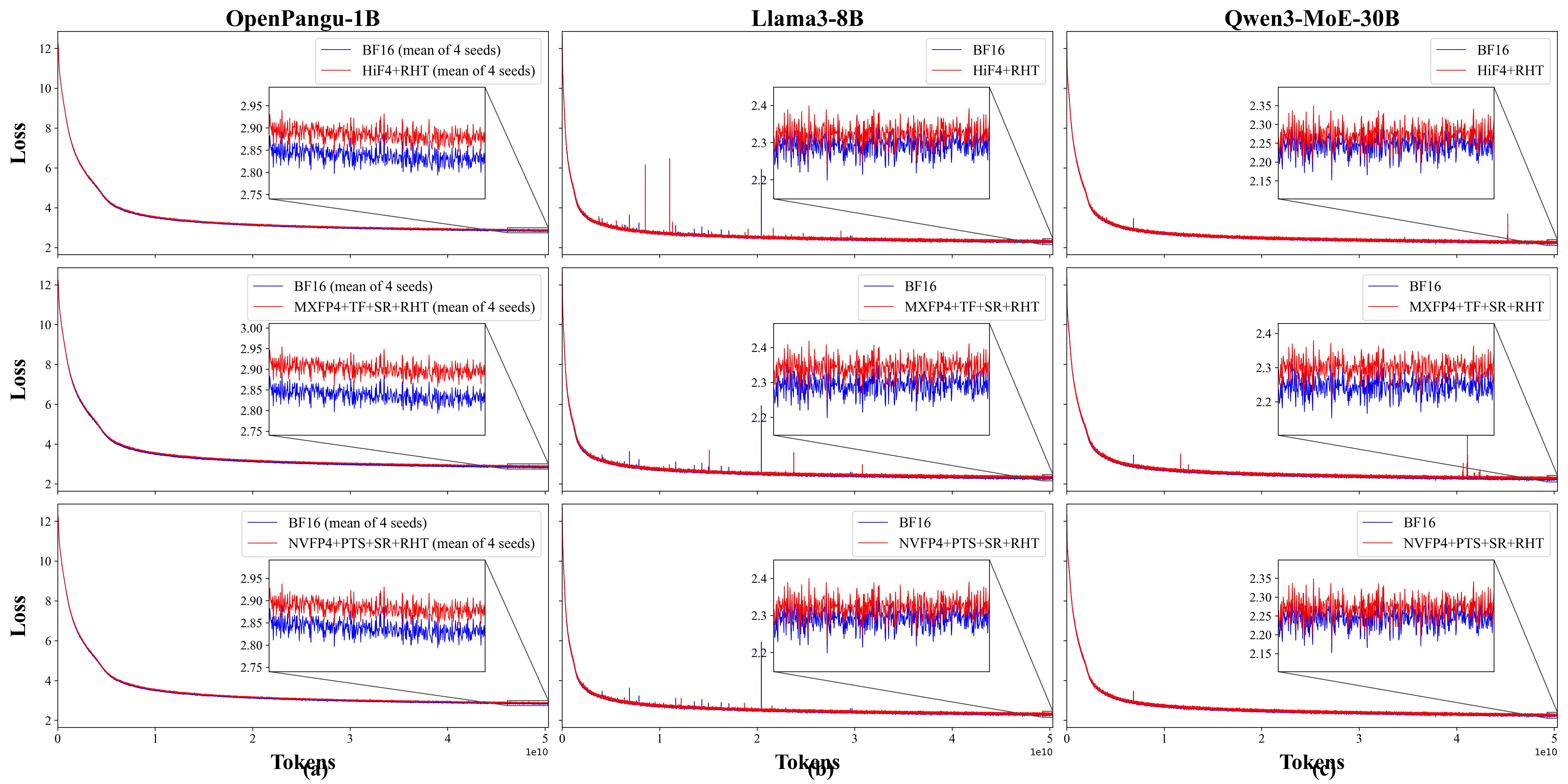}
    \caption{Training loss curves for OpenPangu-1B (a), Llama3-8B (b) and Qwen3-MoE-30B (c) over 50B tokens, each FP4 scheme against BF16. Top: HiF4+RHT. Middle: MXFP4+TF+SR+RHT. Bottom: NVFP4+TS+SR+RHT. OpenPangu-1B is averaged across 4 seeds; insets show the late-training region where the gap is most visible. HiF4 tracks BF16 at least as closely as MXFP4 and NVFP4 on all three models despite using fewer stabilizers.}
    \label{fig:fp4-curves}
\end{figure}

\begin{table}[!htbp]
    \centering
    \caption{Relative loss error with respect to the BF16 baseline across three models trained on 50B tokens, ordered by active parameter count. Each format uses its own best configuration (RHT for HiF4; TF+SR+RHT for MXFP4; PTS+SR+RHT for NVFP4, see Section~\ref{sec:ablation}). OpenPangu-1B is averaged across 4 seeds; the larger models are single runs without error bars.}
    \begin{tabular}{l|c|ccc}
        \toprule
        Model & Activate Parameters & MXFP4 & NVFP4 & HiFloat4  \\
        \midrule
        OpenPangu-1B & 1B & 1.75\% & 1.26\% & \textbf{1.25}\%\\
        Llama3-8B & 8B & 1.44\% & 0.99\% & \textbf{0.85}\%\\
        Qwen3-MoE-30B & 4.6B & 1.55\% & 0.99\% & \textbf{0.88}\%\\
        \bottomrule
    \end{tabular}
    \label{tab:relative-error}
\end{table}

\subsection{Downstream Performance}
We evaluate the three pretrained models on tasks spanning science knowledge, reading comprehension, commonsense reasoning, and language modeling. Specifically, we report 5-shot accuracy on AI2 Reasoning Challenge--Easy (ARC-E)~\citep{arc-e}, HellaSwag~\citep{zellers2019hellaswag}, PIQA~\citep{bisk2020piqa}, SciQ~\citep{welbl2017crowdsourcing}, and BoolQ~\citep{clark2019boolq}, and and WinoGrande~\citep{sakaguchi2021winogrande}, plus WikiText perplexity~\citep{merity2016pointer}, comparing each FP4 format against the BF16 baseline. Results are summarized in Table~\ref{tab:downstream}.

The downstream results track the loss results and add little to them. HiF4 has the smallest mean
accuracy gap on both dense models ($-0.71$ and $-0.24$, against $-1.59$/$-0.71$ for MXFP4 and
$-1.41$/$-1.29$ for NVFP4); on Qwen3-MoE-30B the three fall within $0.3$ points of one another, with
NVFP4 nominally ahead, which we do not read as separating them. On WikiText, HiF4 and NVFP4 are
effectively tied and both clearly ahead of MXFP4: the increase over BF16 is at most $+0.71$ for HiF4 and
$+0.69$ for NVFP4 against $+1.03$ for MXFP4, with HiF4 and NVFP4 never differing by more than $0.03$ on
any model. Both of the 4.5-bit formats see their perplexity penalty roughly halve from 1B to 30B while
MXFP4's never falls below $+0.60$, which mirrors Table~\ref{tab:relative-error}.

BoolQ and ARC-C are by a wide margin the noisiest benchmarks in the suite, and both move by
several points in either direction across formats and models. On Llama3-8B, BoolQ gains $+4.96$ for
HiF4 and $+4.04$ for MXFP4, while on OpenPangu-1B and Qwen3-MoE-30B it loses $-6.19$ and $-7.23$ for
NVFP4. ARC-C falls for every format on both dense models, then rises on Qwen3-MoE-30B for MXFP4
($+1.57$) and NVFP4 ($+2.28$) but not for HiF4 ($+0.14$). Since perplexity rises for every FP4 format
on every model, we read the positive swings on both tasks as evaluation noise rather than genuine
modeling improvement: a benchmark that moves down at two scales and up at the third, with no
consistent ordering among the formats, is reporting seed variance. The negative swings under NVFP4 on
BoolQ are one-sided, however, and that one task accounts for much of its mean accuracy gap; we flag
it as a caveat rather than a conclusion, since one binary-choice benchmark cannot separate a
format-level effect from seed variance.
\newcommand{\R}[1]{{\color{red}#1}}   
\newcommand{\B}[1]{{\color{blue}#1}}  

\begin{table}[h]
\centering
\small
\setlength{\tabcolsep}{4pt}
\caption{Comparison of HiF4, NVFP4, and MXFP4 against the BF16 baseline across three
pretrained models. BF16 rows report absolute 5-shot accuracy (\%), averaged over three
seeds, and WikiText perplexity (PPL; lower is better, single run). FP4 rows report the
\emph{change} relative to the BF16 row of the same model, with {\color{red}red}
indicating degradation and {\color{blue}blue} improvement; absolute FP4 values are
recovered by adding the two. Mean is computed over the seven accuracy benchmarks only.
The best FP4 result per model is shown in bold. HiF4 yields a smaller mean accuracy gap
and lower perplexity increase than MXFP4 across all three models; NVFP4 attains the
lowest perplexity of the three FP4 formats but degrades sharply on BoolQ.}
\label{tab:downstream}

\begin{tabular}{llccccccc|c|c}
\toprule
Model & Quant & ARC-E & ARC-C & HellaS & PIQA & SciQ & BoolQ & WinoG & Mean & PPL \\
\midrule
\multirow{4}{*}{OpenPangu-1B}
 & BF16  & 47.47 & 26.42 & 35.12 & 65.23 & 80.97 & 60.94 & 51.70 & 52.55 & 28.87 \\
 & MXFP4 & \R{-1.53} & \R{-0.66} & \R{-1.11} & \R{-0.56} & \R{-2.13} & \R{-3.39} & \R{-1.79} & \R{-1.59} & \R{+1.03} \\ 
 & NVFP4 & \R{-1.65} & \R{-0.40} & \R{-0.75} & \B{+0.11} & \R{-0.43} & \R{-6.19} & \R{-0.58} & \R{-1.41} & \textbf{\R{+0.69}} \\
 & HiF4  & \B{+0.28} & \R{-0.52} & \R{-0.88} & \R{-1.18} & \R{-2.47} & \R{-0.75} & \B{+0.53} & \textbf{\R{-0.71}} & \R{+0.71} \\
\midrule
\multirow{4}{*}{Llama3-8B}
 & BF16  & 67.32 & 36.89 & 59.03 & 74.43 & 93.47 & 57.80 & 58.01 & 63.85 & 15.19 \\
 & MXFP4 & \R{-1.08} & \R{-2.78} & \R{-1.79} & \R{-0.47} & \B{+0.07} & \B{+4.04} & \R{-2.97} & \R{-0.71} & \R{+0.60} \\
 & NVFP4 & \R{-1.81} & \R{-2.41} & \R{-0.78} & \B{+0.73} & \R{-0.40} & \R{-1.90} & \R{-2.47} & \R{-1.29} & \textbf{\R{+0.36}} \\
 & HiF4  & \R{-1.72} & \R{-1.62} & \R{-0.83} & \R{-0.74} & \B{+0.17} & \B{+4.96} & \R{-1.87} & \textbf{\R{-0.24}} & \R{+0.38} \\
\midrule
\multirow{4}{*}{Qwen3-MoE-30B}
 & BF16  & 68.81 & 36.66 & 61.16 & 75.75 & 93.97 & 59.80 & 58.33 & 64.93 & 14.45 \\
 & MXFP4 & \R{-0.92} & \B{+1.57} & \R{-1.99} & \R{-0.56} & \R{-0.24} & \R{-3.55} & \R{-2.79} & \R{-1.21} & \R{+0.78} \\
 & NVFP4 & \B{+0.62} & \B{+2.28} & \R{-0.76} & \B{+0.02} & \B{+0.36} & \R{-7.23} & \R{-1.63} & \textbf{\R{-0.91}} & \textbf{\R{+0.31}} \\
 & HiF4  & \R{-1.65} & \B{+0.14} & \R{-1.08} & \R{-0.47} & \R{-0.24} & \R{-2.36} & \R{-2.61} & \R{-1.18} & \R{+0.34} \\
\bottomrule
\end{tabular}
\end{table}

\subsection{Sensitivity Analysis}
\label{sec:ablation}
We ablate how sensitive the three formats are to the stabilization techniques available to each: SR, RHT, TF scaling for MXFP4, and PTS for NVFP4. TF scaling does not apply to HiF4 or NVFP4, whose scaling designs avoid the truncation problem it fixes, and PTS does not apply to MXFP4 or HiF4, whose scalers already span the required range. Figure~\ref{fig:ablation}a reports relative loss vs.\ BF16 for each combination on OpenPangu-1B, and Figure~\ref{fig:ablation}b the key configurations on Qwen3-MoE-30B.

The formats differ much more in how much stabilization they need than in where they end up. HiF4 with nothing applied reaches 1.55\%, already better than MXFP4 with all three of its stabilizers (1.79\%) and better than NVFP4 with its own best pair, PTS and RHT (1.42\%), let alone with PTS alone (2.00\%). Adding RHT takes HiF4 to 1.19\%, the best number we measured for any format. HiF4 reaches it with one technique; the other two need three. 

\begin{wrapfigure}{r}{0.6\textwidth}
  \centering
  \includegraphics[width=0.6\textwidth]{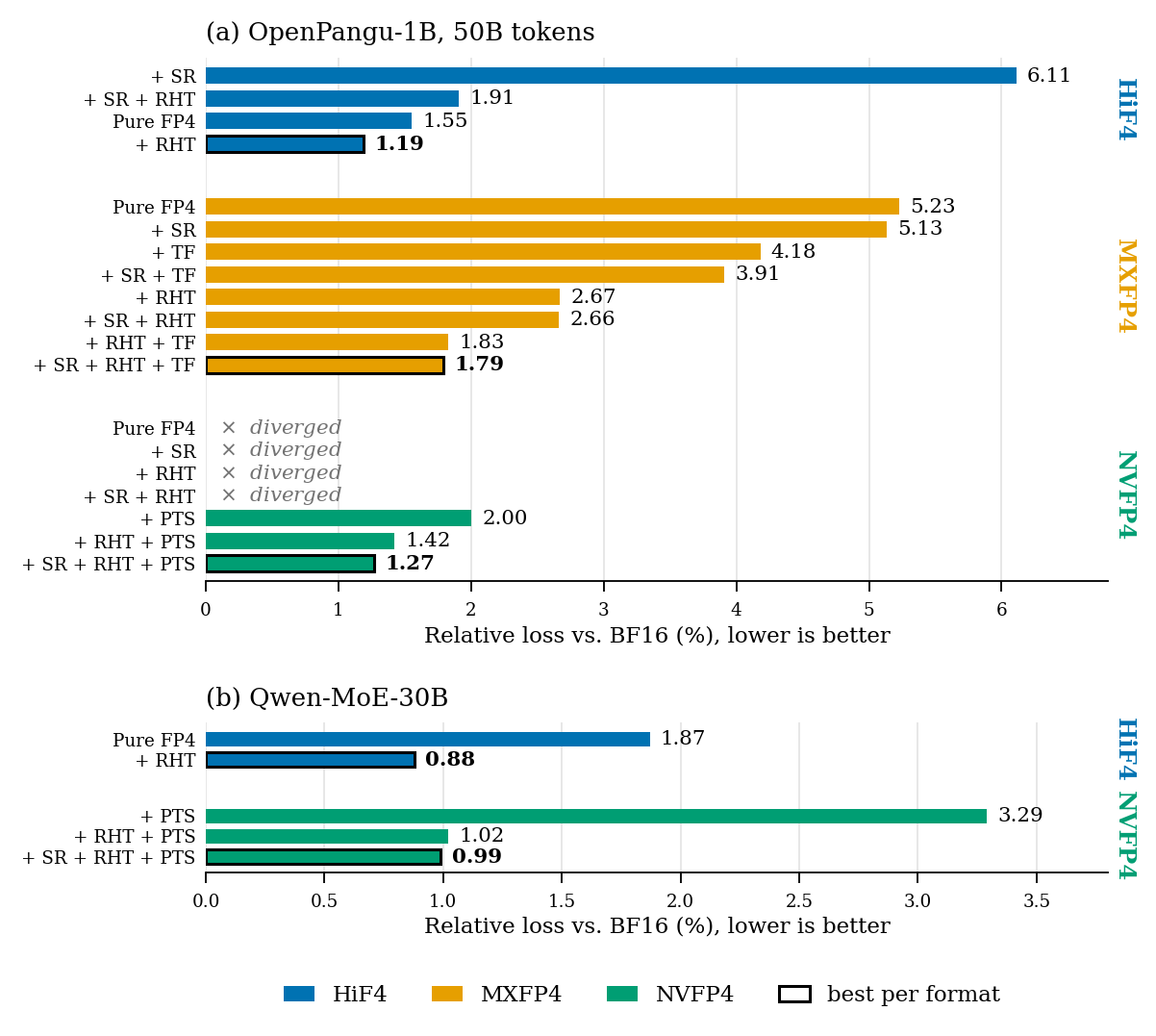}
  \caption{Ablation of FP4 stabilizers on (a) OpenPangu-1B and (b) Qwen-MoE-30B, both trained for 50B tokens. Stabilizers are SR, RHT, TF, and PTS; TF applies only to MXFP4 and PTS only to NVFP4. Bars show loss relative to the BF16 baseline; lower is better. NVFP4 diverges without PTS. The Qwen-MoE-30B runs cover HiF4 and NVFP4 only, testing whether the 1B ordering holds at scale. Best configuration per format is outlined and in bold.}
  \label{fig:ablation}
\end{wrapfigure} 
NVFP4's dependence is not a matter of degree. Of its three techniques, PTS is the one it cannot do without: plain NVFP4 diverges, and so do $+$SR, $+$RHT and $+$SR$+$RHT. This follows from the format's range rather than from rounding. RHT spreads energy across dimensions and SR removes rounding bias, but neither changes which magnitudes a format can hold, and NVFP4 covers 22 binades against HiF4's 69. \citet{luo2026hifloat4} see the same failure at inference time, where direct-cast NVFP4 collapses and per-tensor scaling fixes it. With PTS applied, the usual ordering returns: RHT does most of the remaining work ($2.00 \rightarrow 1.42\%$) and SR adds a little more ($1.42 \rightarrow 1.27\%$). Every one of those three runs in high precision outside the FP4 kernel, so NVFP4's best result carries the full stack.
Figure~\ref{fig:ablation}b shows this is not specific to the 1B model. On Qwen3-MoE-30B, a different architecture at larger scale, HiF4 leads throughout, and HiF4 with no stabilization (1.87\%) is again closer to BF16 than NVFP4 with PTS (3.29\%). The dependence pattern holds as well: RHT buys NVFP4 2.27 points here and 0.58 at 1B, against 0.99 and 0.36 for HiF4. Within each format the picture is just as distinct. For HiF4, RHT is the only technique that helps ($-0.36\%$). SR hurts: it adds 4.56\% on its own, and even with RHT it is worse than RHT alone and worse than plain HiF4. We think this is because HiF4's scaling already keeps the small-magnitude information that SR exists to recover. For MXFP4, RHT does most of the work ($-2.56\%$, the largest single-technique gain), TF contributes $1.05\%$, and SR on its own is near-neutral ($0.10\%$); RHT with TF ($1.83\%$) gets almost all the way to the full recipe ($1.79\%$). So HiF4 is at its best with one stabilizer and gets worse with two, while the other two formats keep improving through three.

\section{Conclusion and Future Directions}
\label{sec:conclusion}
We present a systematic study of end-to-end FP4 pretraining for large language models on Ascend NPUs,
comparing HiFloat4 (HiF4) against MXFP4 and NVFP4 under one fixed recipe. We claim that the hierarchical scaling is enough on its own for stable pretraining, where gradient quantization is the difficulty inference work never meets. HiF4 and NVFP4 end up close on accuracy -- HiF4 still better. What separates them is
how much high-precision help each needs first: HiF4 is competitive with none, while MXFP4 and NVFP4 each need three mechanisms, and NVFP4 does not converge at all without PTS. A stabilization recipe therefore belongs to a format and does not port between formats --- SR is necessary for MXFP4, harmful for HiF4, and on its own not enough for NVFP4.

\textbf{Limitations.}
Our comparison is fixed at 50B tokens, and nothing here speaks to convergence over long horizons. \citet{abecassis2025pretraining} report instabilities appearing late in much longer runs and caution that small-scale conclusions may not transfer; since their gaps also widen with tokens, our absolute numbers likely understate a long run. All FP4 arithmetic is emulated, so we evaluate numerical behaviour only; for HiF4 no silicon exists yet. Only OpenPangu-1B is multi-seed, and the 0.08--0.14 point HiF4--NVFP4 margins are not separable from
run-to-run variation on this evidence, which is why the paper rests on the stabilization results instead. We omit 2D weight quantization, an untested deviation from the NVIDIA recipe applied equally to all three formats, and we do not study spending MXFP4's 0.25-bit saving on higher-precision sensitive layers --- a budget HiF4 could spend the same way. Our ablation also establishes the
stabilization ordering only as an early-horizon property, which is where divergence happens.

\textbf{Future Directions.}
Extending FP4 to post-training, especially RL fine-tuning, is unexplored and high-impact, as are long-context and multimodal training. The long-horizon study this paper cannot provide becomes affordable, and far more useful, once native HiF4 silicon exists: convergence and speedup could then be measured in the same run.

\bibliographystyle{plainnat}
\bibliography{ref}

@article{rouhani2023microscaling,
  title={Microscaling data formats for deep learning},
  author={Rouhani, Bita Darvish and Zhao, Ritchie and More, Ankit and Hall, Mathew and Khodamoradi, Alireza and Deng, Summer and Choudhary, Dhruv and Cornea, Marius and Dellinger, Eric and Denolf, Kristof and others},
  journal={arXiv preprint arXiv:2310.10537},
  year={2023}
}

@article{chen2025oscillation,
  title={Oscillation-reduced mxfp4 training for vision transformers},
  author={Chen, Yuxiang and Xi, Haocheng and Zhu, Jun and Chen, Jianfei},
  journal={arXiv preprint arXiv:2502.20853},
  year={2025}
}

@article{croci2022stochastic,
  title={Stochastic rounding: implementation, error analysis and applications},
  author={Croci, Matteo and Fasi, Massimiliano and Higham, Nicholas J and Mary, Theo and Mikaitis, Mantas},
  journal={Royal Society Open Science},
  volume={9},
  number={3},
  year={2022},
  publisher={The Royal Society}
}

@article{luo2026hifloat4,
  title={HiFloat4 Format for Language Model Inference},
  author={Luo, Yuanyong and Huang, Jing and Cheng, Yu and Yu, Ziwei and Zhang, Kaihua and Hong, Kehong and Ma, Xinda and Wang, Xin and Tong, Anping and Hu, Guipeng and others},
  journal={arXiv preprint arXiv:2602.11287},
  year={2026}
}

@techreport{rouhani2023mxfp4,
  author = {Rouhani, Bita Darvish and Garegrat, Nitin and Savell, Tom and More, Ankit and Han, Kyung-Nam and Zhao, Mathew and Hall, Ritchie and Klar, Jasmine and Chung, Eric and Yu, Yuan and Schulte, Michael and Wittig, Ralph and Bratt, Ian and Stephens, Nigel and Milanovic, Jelena and Brothers, John and Dubey, Pradeep and Cornea, Marius and Heinecke, Alexander and Rodriguez, Andres and Langhammer, Martin and Deng, Summer and Naumov, Maxim and Micikevicius, Paulius and Siu, Michael and Verrilli, Colin},
  title = {OCP Microscaling (MX) Specification},
  institution = {Open Compute Project},
  year = {2023}
}

@article{cook2025four,
  title={Four Over Six: More Accurate NVFP4 Quantization with Adaptive Block Scaling},
  author={Cook, Jack and Guo, Junxian and Xiao, Guangxuan and Lin, Yujun and Han, Song},
  journal={arXiv preprint arXiv:2512.02010},
  year={2025}
}

@article{zhang2025accurate,
  title={Accurate int8 training through dynamic block-level fallback},
  author={Zhang, Pengle and Wei, Jia and Zhang, Jintao and Zhu, Jun and Chen, Jianfei},
  journal={arXiv preprint arXiv:2503.08040},
  year={2025}
}

@article{chmiel2025fp4,
  title={Fp4 all the way: Fully quantized training of llms},
  author={Chmiel, Brian and Fishman, Maxim and Banner, Ron and Soudry, Daniel},
  journal={arXiv preprint arXiv:2505.19115},
  year={2025}
}

@article{hu2025elucidating,
  title={Elucidating the Design Space of FP4 training},
  author={Hu, Robert and Luschi, Carlo and Balanca, Paul},
  journal={arXiv preprint arXiv:2509.17791},
  year={2025}
}

@article{chen2025int,
  title={INT vs FP: A Comprehensive Study of Fine-Grained Low-bit Quantization Formats},
  author={Chen, Mengzhao and Wu, Meng and Jin, Hui and Yuan, Zhihang and Liu, Jing and Zhang, Chaoyi and Li, Yunshui and Huang, Jie and Ma, Jin and Xue, Zeyue and others},
  journal={arXiv preprint arXiv:2510.25602},
  year={2025}
}

@article{zhang2026practical,
  title={Practical FP4 Training for Large-Scale MoE Models on Hopper GPUs},
  author={Zhang, Wuyue and Huang, Chongdong and You, Chunbo and Gu, Cheng and Wang, Fengjuan and Sun, Mou},
  journal={arXiv preprint arXiv:2603.02731},
  year={2026}
}

@article{chen2025power,
  title={The power of negative zero: Datatype customization for quantized large language models},
  author={Chen, Yuzong and Dai, Xilai and Chang, Chi-chih and Akhauri, Yash and Abdelfattah, Mohamed S},
  journal={arXiv preprint arXiv:2501.04052},
  year={2025}
}

@misc{yan2026scalabletrainingmixtureofexpertsmodels,
      title={Scalable Training of Mixture-of-Experts Models with Megatron Core}, 
      author={Zijie Yan and Hongxiao Bai and Xin Yao and Dennis Liu and Tong Liu and Hongbin Liu and Pingtian Li and Evan Wu and Shiqing Fan and Li Tao and Robin Zhang and Yuzhong Wang and Shifang Xu and Jack Chang and Xuwen Chen and Kunlun Li and Yan Bai and Gao Deng and Nan Zheng and Vijay Anand Korthikanti and Abhinav Khattar and Ethan He and Soham Govande and Sangkug Lym and Zhongbo Zhu and Qi Zhang and Haochen Yuan and Xiaowei Ren and Deyu Fu and Tailai Ma and Shunkang Zhang and Jiang Shao and Ray Wang and Vasudevan Rengasamy and Rachit Garg and Santosh Bhavani and Xipeng Li and Chandler Zhou and David Wu and Yingcan Wei and Ashwath Aithal and Michael Andersch and Mohammad Shoeybi and Jiajie Yao and June Yang},
      year={2026},
      eprint={2603.07685},
      archivePrefix={arXiv},
      primaryClass={cs.DC},
      url={https://arxiv.org/abs/2603.07685}, 
}

@article{cao2025metis,
  title={Metis: Training LLMs with FP4 Quantization},
  author={Cao, Hengjie and Chen, Mengyi and Yang, Yifeng and Huang, Ruijun and Dong, Fang and Zhou, Jixian and Chen, Anrui and Dong, Mingzhi and Wang, Yujiang and Hou, Jinlong and others},
  journal={arXiv preprint arXiv:2509.00404},
  year={2025}
}

@article{wang2025optimizing,
  title={Optimizing large language model training using fp4 quantization},
  author={Wang, Ruizhe and Gong, Yeyun and Liu, Xiao and Zhao, Guoshuai and Yang, Ziyue and Guo, Baining and Zha, Zhengjun and Cheng, Peng},
  journal={arXiv preprint arXiv:2501.17116},
  year={2025}
}

@article{abecassis2025pretraining,
  title={Pretraining large language models with nvfp4},
  author={Abecassis, Felix and Agrusa, Anjulie and Ahn, Dong and Alben, Jonah and Alborghetti, Stefania and Andersch, Michael and Arayandi, Sivakumar and Bjorlin, Alexis and Blakeman, Aaron and Briones, Evan and others},
  journal={arXiv preprint arXiv:2509.25149},
  year={2025}
}

@article{panferov2026quartet,
  title={Quartet II: Accurate LLM Pre-Training in NVFP4 by Improved Unbiased Gradient Estimation},
  author={Panferov, Andrei and Schultheis, Erik and Tabesh, Soroush and Alistarh, Dan},
  journal={arXiv preprint arXiv:2601.22813},
  year={2026}
}

@article{castro2025quartet,
  title={Quartet: Native fp4 training can be optimal for large language models},
  author={Castro, Roberto L and Panferov, Andrei and Tabesh, Soroush and Sieberling, Oliver and Chen, Jiale and Nikdan, Mahdi and Ashkboos, Saleh and Alistarh, Dan},
  journal={arXiv preprint arXiv:2505.14669},
  year={2025}
}

@article{chmiel2021accurate,
  title={Accurate neural training with 4-bit matrix multiplications at standard formats},
  author={Chmiel, Brian and Banner, Ron and Hoffer, Elad and Yaacov, Hilla Ben and Soudry, Daniel},
  journal={arXiv preprint arXiv:2112.10769},
  year={2021}
}

@article{zhou2025towards,
  title={Towards efficient pre-training: Exploring fp4 precision in large language models},
  author={Zhou, Jiecheng and Tang, Ding and Fu, Rong and Hu, Boni and Xu, Haoran and Wang, Yi and Pei, Zhilin and Su, Zhongling and Liu, Liang and Zhang, Xingcheng and others},
  journal={arXiv preprint arXiv:2502.11458},
  year={2025}
}

@article{tseng2025training,
  title={Training llms with mxfp4},
  author={Tseng, Albert and Yu, Tao and Park, Youngsuk},
  journal={arXiv preprint arXiv:2502.20586},
  year={2025}
}

@article{hu2026m2xfp,
  title={M2XFP: A Metadata-Augmented Microscaling Data Format for Efficient Low-bit Quantization},
  author={Hu, Weiming and Zhang, Zihan and Zhang, Haoyan and Zhang, Chen and Guo, Cong and Feng, Yu and Hu, Tianchi and Li, Guanglin and Hu, Guipeng and Wang, Junsong and others},
  journal={arXiv e-prints},
  pages={arXiv--2601},
  year={2026}
}

@article{kaplan2020scaling,
  title={Scaling laws for neural language models},
  author={Kaplan, Jared and McCandlish, Sam and Henighan, Tom and Brown, Tom B and Chess, Benjamin and Child, Rewon and Gray, Scott and Radford, Alec and Wu, Jeffrey and Amodei, Dario},
  journal={arXiv preprint arXiv:2001.08361},
  year={2020}
}

@article{micikevicius2022fp8,
  title={Fp8 formats for deep learning},
  author={Micikevicius, Paulius and Stosic, Dusan and Burgess, Neil and Cornea, Marius and Dubey, Pradeep and Grisenthwaite, Richard and Ha, Sangwon and Heinecke, Alexander and Judd, Patrick and Kamalu, John and others},
  journal={arXiv preprint arXiv:2209.05433},
  year={2022}
}

@article{liu2024deepseek,
  title={Deepseek-v3 technical report},
  author={Liu, Aixin and Feng, Bei and Xue, Bing and Wang, Bingxuan and Wu, Bochao and Lu, Chengda and Zhao, Chenggang and Deng, Chengqi and Zhang, Chenyu and Ruan, Chong and others},
  journal={arXiv preprint arXiv:2412.19437},
  year={2024}
}

@Inbook{huawei2023huawei,
title="Huawei Atlas AI Computing Solution",
bookTitle="Artificial Intelligence Technology",
year="2023",
publisher="Springer Nature Singapore",
address="Singapore",
pages="163--219",
isbn="978-981-19-2879-6",
doi="10.1007/978-981-19-2879-6_6",
url="https://doi.org/10.1007/978-981-19-2879-6_6"
}

@article{arc-e,
  title={Think you have solved question answering? try arc, the ai2 reasoning challenge},
  author={Clark, Peter and Cowhey, Isaac and Etzioni, Oren and Khot, Tushar and Sabharwal, Ashish and Schoenick, Carissa and Tafjord, Oyvind},
  journal={arXiv preprint arXiv:1803.05457},
  year={2018}
}

@inproceedings{zellers2019hellaswag,
  title={Hellaswag: Can a machine really finish your sentence?},
  author={Zellers, Rowan and Holtzman, Ari and Bisk, Yonatan and Farhadi, Ali and Choi, Yejin},
  booktitle={Proceedings of the 57th annual meeting of the association for computational linguistics},
  pages={4791--4800},
  year={2019}
}

@inproceedings{bisk2020piqa,
  title={Piqa: Reasoning about physical commonsense in natural language},
  author={Bisk, Yonatan and Zellers, Rowan and Gao, Jianfeng and Choi, Yejin and others},
  booktitle={Proceedings of the AAAI conference on artificial intelligence},
  volume={34},
  number={05},
  pages={7432--7439},
  year={2020}
}

@inproceedings{welbl2017crowdsourcing,
  title={Crowdsourcing multiple choice science questions},
  author={Welbl, Johannes and Liu, Nelson F and Gardner, Matt},
  booktitle={Proceedings of the 3rd Workshop on Noisy User-generated Text},
  pages={94--106},
  year={2017}
}

@inproceedings{clark2019boolq,
  title={Boolq: Exploring the surprising difficulty of natural yes/no questions},
  author={Clark, Christopher and Lee, Kenton and Chang, Ming-Wei and Kwiatkowski, Tom and Collins, Michael and Toutanova, Kristina},
  booktitle={Proceedings of the 2019 conference of the north American chapter of the association for computational linguistics: Human language technologies, volume 1 (long and short papers)},
  pages={2924--2936},
  year={2019}
}

@article{merity2016pointer,
  title={Pointer sentinel mixture models},
  author={Merity, Stephen and Xiong, Caiming and Bradbury, James and Socher, Richard},
  journal={arXiv preprint arXiv:1609.07843},
  year={2016}
}

@article{sakaguchi2021winogrande,
  title={WinoGrande: An adversarial Winograd schema challenge at scale},
  author={Sakaguchi, Keisuke and Le Bras, Ronan and Bhagavatula, Chandra and Choi, Yejin},
  journal={Communications of the ACM},
  volume={64},
  number={9},
  pages={99--106},
  year={2021}
}

\appendix
\section{Training Configuration}
\label{sec:config}
\begin{table}[h]
\centering
\caption{Training configurations for all experiments. }
\label{tab:training_config}
\small
\begin{tabular}{l c c c}
\toprule
\textbf{Configuration} & \textbf{OpenPangu-1B} & \textbf{Llama3-8B} & \textbf{Qwen3-MoE-30B} \\
\midrule
\multicolumn{3}{l}{\textit{Model \& Data}} \\
Training Tokens & 50B &  50B & 50B\\
Sequence Length & 4K & 4K & 4K\\

\midrule
\multicolumn{3}{l}{\textit{Optimization}} \\
Optimizer & AdamW & AdamW & AdamW\\
Start Learning Rate & $10^{-4}$ & $10^{-4}$ & $10^{-4}$ \\
End Learning Rate & $10^{-5}$ & $10^{-5}$ & $10^{-5}$ \\
LR Schedule & Cosine Decay & Cosine Decay & Cosine Decay \\
Warmup Steps & 480 & 2K & 2K \\
Global Batch Size & 2048 & 512 & 512 \\

\bottomrule
\end{tabular}
\end{table}

\section{Low-Precision Formats}
\label{sec:low-precision-formats}
\subsection{MXFP4}
MXFP4 \citep{rouhani2023microscaling, rouhani2023mxfp4} is a 4-bit floating-point format proposed within the Open Compute Project (OCP) framework. It adopts a block-wise microscaling design in which each tensor is partitioned into blocks of 32 elements. Within each block, individual values are encoded using a private E2M1 representation (2 exponent bits and 1 mantissa bit), while a shared E8M0 scaling factor provides a common block-level scale.
The E2M1 format supports representable values in $\pm\{0, 0.5, 1, 1.5, 2, 3, 4, 6\}$, whereas the shared E8M0 scaler spans a dynamic range of $\left[2^{-127}, 2^{127}\right]$. The effective representable range of each block is therefore determined by the product of the shared E8M0 scale and the local E2M1 values, enabling substantial dynamic range expansion while maintaining compact 4-bit storage. Figure \ref{fig:mxfp4} illustrates MXFP4 architecture.
To mitigate information loss from out-of-range values exceeding $\pm6$, we adopt a truncation-free scaling strategy similar to TetraJet \citep{chen2025oscillation}. Instead of clipping large magnitudes to the representable E2M1 limits, the block-level scaling factor is adjusted to ensure that all values remain within range, thereby avoiding severe distortion introduced by hard truncation.

\subsection{HiFloat4}
HiFloat4 (HiF4) is a hardware-oriented FP4 format designed for Huawei Ascend NPUs, featuring a three-level hierarchical scaling scheme to improve both precision and dynamic range while maintaining 4-bit storage per value.
At the first level, HiF4 employs a shared unsigned 8-bit E6M2 floating-point scale, allocating 6 bits to the exponent (bias 48) and 2 bits to the mantissa with an implicit leading 1. The value is represented as $X_{E6M2} = 2^E\times 1.M$. This level provides a coarse global scaling factor for each block.
The second and third levels introduce fine-grained scaling through 1-bit micro-exponents, denoted as $E1_8$ and $E1_{16}$, corresponding to 8-way and 16-way hierarchical partitioning, respectively. 
In this structure, the level-1 E6M2 factor serves as the global base scale shared across the block. Each level-2 micro-exponent modulates a subset of the block and is further refined by two adjacent level-3 micro-exponents. Each level-3 micro-exponent governs four contiguous 4-bit data elements within its local group, enabling localized dynamic range adaptation. The metadata for all three scaling levels occupies 32 bits per block of 64 elements, resulting in an amortized overhead of 0.5 bits per value.
The 4-bit data elements are encoded in S1P2 format, where S denotes the sign bit, “1” represents the integer bit preceding the binary point, and “2” corresponds to two fractional bits. This representation is equivalent to an E1M2 floating-point format. Figure \ref{fig:hif4} illustrates the HiF4 architecture.
For a detailed architectural description and design rationale, we refer the reader to \citep{luo2026hifloat4}.

\section{Stabilization Techniques for Low-Precision Training}
\label{sec:stabilization}
\subsection{Unbiased Quantized Training with Stochastic Rounding}
Training with low-bit formats requires mapping scaled values to a small set of representable points. The standard choice, NR, deterministically selects the nearest representable value and thereby introduces a systematic quantization bias. This bias is particularly harmful for gradients, where it accumulates across iterations and can degrade optimization dynamics. To mitigate this, we adopt SR, which introduces controlled randomness prior to rounding. Concretely, for a value $x$, we sample $\Delta \sim \mathcal{U}(-\delta, \delta)$ and compute $\hat{x} = \mathrm{Round}(x + \Delta)$, where $\mathrm{Round}(\cdot)$ denotes rounding to the nearest representable value in the target 4-bit format, and $\delta = 0.5$ controls the noise magnitude. By construction, SR yields unbiased quantization in expectation, eliminating the systematic error inherent to NR.

\subsection{Low-Variance Quantized Training with Random Hadamard Transform}
To further stabilize low-precision quantization, we apply an RHT to the input tensors prior to quantization. The transform redistributes energy more uniformly across dimensions, suppressing outliers and reducing the per-coordinate dynamic range. This mitigates underflow of small values in the presence of large outliers and, when combined with SR, reduces the variance of the noise that SR injects.
Formally, given an input tensor $X \in \mathbb{R}^{j \times k}$, the transform is defined as $\tilde{X} \leftarrow H S X$, where $S = \mathrm{diag}(d_1, \dots, d_k) \in \mathbb{R}^{k \times k}$ is a random sign matrix with $d_i \overset{\text{i.i.d.}}{\sim} \mathrm{Uniform}\{\pm 1\}$, and $H \in \mathbb{R}^{k \times k}$ is the normalized Hadamard matrix, defined recursively as
\begin{equation}
    H_n = \frac{1}{\sqrt{2}}
    \begin{bmatrix}
        H_{n-1} & H_{n-1} \\
        H_{n-1} & -H_{n-1}
    \end{bmatrix},
    \qquad H_1 = [1].
\end{equation}
Since $H$ and $S$ are both orthogonal, we have $HS(HS)^\top = I$. Consequently, inserting RHT into a GEMM preserves the exact matrix product $A^\top B = (HSA)^\top (HSB)$, ensuring mathematical equivalence in full precision while improving quantization robustness under low-precision arithmetic.

\section{Analyzing Computation Complexity of Metis and Quartet II}
\label{sec:compute-analysis}
\subsection{NVIDIA's recipe}
NVIDIA stabilizes NVFP4 backward passes with a single tool: an RHT applied along the contracted dimension of each backward GeMM. We assume that the rotation uses a Walsh-Hadamard structure, applying it to a tensor of $T$ elements costs $T·\log(r)$, where r is the rotation block size. Each backward GeMM has two operands that need rotating, so the total extra cost per backward pass is roughly (matrix perimeter) × log r. The forward pass adds nothing — it relies on 2D weight quantization to keep weights reusable across forward and backward, avoiding any extra rotation work.

\subsection{Quartet II}
Quartet II uses the same basic tool — Hadamard rotations on the backward inner dimension — but adds an unbiasedness correction (MS-EDEN) on top of round-to-nearest quantization, and re-quantizes weights on the backward pass instead of reusing them. At the asymptotic level, the rotation cost is identical to NVIDIA's: (matrix perimeter) $\times \log(r)$ per backward GeMM. The MS-EDEN correction adds a small inner-product computation per rotation block, but this is sub-leading and gets absorbed into the rotation cost. On the forward pass, Quartet II's 4/6 scale selection roughly doubles per-element quantization work, but quantization is $\Theta(matrix size)$ and dwarfed by the GeMM, so this doesn't show up in the asymptotic table either. 

\subsection{Metis}
Metis takes a fundamentally different approach and pays a fundamentally different price. Instead of rotating to smooth distributions, it decomposes matrices into a low-rank "dominant" piece and a full-rank "residual" piece, then quantizes each separately. This adds three distinct sources of overhead:

\textbf{Forward path through low-rank factors.} The forward pass now has two routes: a baseline-cost route through the residual matrix, and an extra low-rank route through the rank-k factors. The low-rank route costs rank × (matrix perimeter) — concretely $Mk(K+N)$ for input X and rank-k factors of W.

\textbf{Per-step gradient decomposition.} This is the term with no analog elsewhere. Every backward pass, Metis freshly decomposes the gradient matrix D via randomized SVD, which costs roughly $MNj$ where $j$ is the gradient rank. Notice that this term scales with the product $MN$ of two large dimensions — unlike RHT, which only scales with the sum of dimensions times $\log(r)$. This is the dominant cost in Metis and the conceptual reason its overhead is heavier than the rotation-based methods.

\textbf{Mixed contractions.} When the decomposed gradient meets the decomposed weights in computing parameter updates, the resulting chains of small matrix products cost roughly jk × (sum of all three matrix dimensions).
The rank parameters k and j are tunable, which is both a feature and a complication. The paper sets them to 1\% of matrix rank for FP8 (where Metis is essentially free) but 50\% of matrix rank for FP4 (where Metis is the headline result). At 50\%, the $MNj$ term alone becomes comparable to the baseline GeMM cost — Metis's stabilization adds roughly 50\% on top of the original compute, versus a few percent for the rotation-based methods. This is the price of treating anisotropy structurally rather than smoothing it away.

\end{document}